\documentclass{article}

\PassOptionsToPackage{numbers, compress}{natbib}

\usepackage[final]{nips_2018}

\usepackage[utf8]{inputenc} 
\usepackage[T1]{fontenc}    
\usepackage{hyperref}       
\usepackage{url}            
\usepackage{booktabs}       
\usepackage{amsfonts}       
\usepackage{nicefrac}       
\usepackage{microtype}      
\usepackage{xcolor}

\usepackage{graphicx}
\usepackage{subfig}

\title{Web Applicable Computer-aided Diagnosis of Glaucoma Using Deep Learning}

\makeatletter
\def\blfootnote{\xdef\@thefnmark{}\@footnotetext}
\makeatother

\author{
  Mijung Kim\textsuperscript{*\dag} \qquad Olivier Janssens\textsuperscript{*} \qquad Ho-min Park\textsuperscript{*\dag} \\ \textbf{Jasper Zuallaert\textsuperscript{*\dag}} \qquad \textbf{Sofie Van Hoecke\textsuperscript{*}} \qquad \textbf{Wesley De Neve\textsuperscript{*\dag}} \\ \\
  \textsuperscript{\dag}Center for Biotech Data Science, Ghent University Global Campus \\
  \textsuperscript{*}IDLab, Department of Electronics and Information Systems, Ghent University \\
  \texttt{mijung.kim@ugent.be} \\
}

\begin{document}

\maketitle

\begin{abstract}

Glaucoma is a major eye disease, leading to vision loss in the absence of proper medical treatment. Current diagnosis of glaucoma is performed by ophthalmologists who are often analyzing several types of medical images generated by different types of medical equipment. Capturing and analyzing these medical images is labor-intensive and expensive. In this paper, we present a novel computational approach towards glaucoma diagnosis and localization, only making use of eye fundus images that are analyzed by state-of-the-art deep learning techniques. Specifically, our approach leverages Convolutional Neural Networks (CNNs) and Gradient-weighted Class Activation Mapping (Grad-CAM) for glaucoma diagnosis and localization, respectively. Quantitative and qualitative results, as obtained for a small-sized dataset with no segmentation ground truth, demonstrate that the proposed approach is promising, for instance achieving an accuracy of 0.91$\pm0.02$ and an ROC-AUC score of 0.94 for the diagnosis task. Furthermore, we present a publicly available prototype web application that integrates our predictive model, with the goal of making effective glaucoma diagnosis available to a wide audience.

\end{abstract}

\section{Introduction}
\label{sec:intro}

The revival of convolutional neural networks (CNNs) and the public availability of large-scale datasets like ImageNet~\cite{deng2009imagenet} have led to significant performance improvements in computer vision. Unlike its predecessors, ImageNet contains more than one million natural images belonging to more than thousand classes, contributing substantially to the successful deployment of CNNs\blfootnote{* This work has been submitted to the IEEE for possible publication. Copyright may be transferred without notice, after which this version may no longer be accessible.}. Indeed, ImageNet for instance allows constructing CNN-based predictive models that are more accurate and that better generalize for the tasks of image segmentation and classification. Moreover, CNN-based predictive models have shown to be highly successful in medical image analysis~\cite{ronneberger2015u,heinrich2017deep,gulshan2016development}. For diabetic retinopathy~\cite{gulshan2016development}, these models even managed to obtain a diagnosis accuracy that is as high as the diagnosis accuracy of human experts. However, most deep learning-based approaches towards medical image analysis face two major challenges. First, few medical image datasets are available with a volume sufficient to train a deep neural network from scratch, especially given that, by convention, the deeper the neural networks constructed, the better the predictions (i.e., the higher the classification accuracy). As a result, if not enough data are available, then the neural networks constructed may suffer from overfitting. Second, apart from labels for the purpose of classification, most medical image datasets do not have proper annotations available for the purpose of segmentation, given the time-consuming and error-prone nature of annotation efforts.


In our research, to tackle the two aforementioned challenges, we built a predictive model for computer-aided diagnosis (CAD) and localization of glaucoma, leveraging eye fundus images that are widely used in present-day hospitals, given that these images can be acquired at a relatively low cost. Additionally, based on our CAD model, we developed a novel tool for glaucoma diagnosis that takes the form of a prototype web application\footnote{\url{www.medinoid.org}}. The main contribution of this paper stems from the novelty of our predictive model and its integration into a prototype web application. Indeed, although plenty of research efforts have been presented in the scientific literature on computer-aided diagnosis using fundus images, to the best of our knowledge, no previous research efforts have simultaneously targeted both diagnosis and localization using a dataset with no segmentation ground truth.

The rest of this paper is organized as follows. We explain our approach in Section~\ref{sec:model}. We subsequently discuss our experimental setup and results in Section~\ref{sec:experiment}. Finally, we conclude our research in Section~\ref{sec:conclusion}. \textit{Note that some of the research activities described in this paper have also been presented at The Second International Workshop on Deep Learning in Bioinformatics, Biomedicine, and Healthcare Informatics (DLB2H 2018).}

\section{Architecture}
\label{sec:model}

\textbf{Convolutional Neural Network.} Introduced by Simonyan and Zisserman at Oxford's Visual Geometry Group (VGG), VGG networks~\cite{simonyan2014very} are currently some of the most widely used CNN architectures. Among these different architectures, VGG16 and VGG19 are the most popular ones, thanks to their high classification accuracy. In our research, we decided to work with VGG16 over VGG19, given that the difference in effectiveness between the two is not significant, while VGG16 has less parameters to learn than VGG19. However, as mentioned in the introductory section, VGG16 may suffer from overfitting when directly applying this architecture to our eye fundus images. As such, to mitigate overfitting and to increase training effectiveness and efficiency, we applied a number of changes. To facilitate better model regularization, we introduced two techniques. First, we added dropout layers (ratio: 0.5) at the end of each convolutional block. Second, next to the use of default data augmentation techniques (that is, random cropping and random horizontal flipping), we also incorporated data augmentation techniques such as random brightness (using sigma $\pm0.8$). Furthermore, we adopted the adaptive moment estimation (ADAM) optimizer~\cite{kingma2014adam} instead of Stochastic Gradient Descent (SGD), allowing us to handle multiple drawbacks of SGD more efficiently, like a slow convergence and a high variance in parameter updates. Lastly, we used a pre-trained VGG16 model, so to be able to avoid a time-consuming weight initialization process and to considerably reduce the training time~\cite{torrey2010transfer}. From an architectural point-of-view, we also applied a few more changes to the vanilla VGG16 network. First, we used fully convolutional layers instead of fully connected layers. Second, in order to meet our classification needs, the number of channels in the last fully convolutional layer was reduced from 1000 to 2. 

\textbf{Grad-CAM.} Most sets of medical images do not have a ground truth (that is, golden standard annotations) for supervised segmentation or localization. As a result, localization needs to be treated as an unsupervised learning problem. A couple of deep learning-based approaches are frequently used to segment images in an unsupervised fashion. One such approach is Class Activation Mapping (CAM), as proposed by~\cite{zhou2016learning}. CAM allows visualizing what CNNs look at when they classify an input image, thus making it possible to visually understand why a CNN chooses one class over another. More recently, another algorithm was published in~\cite{selvaraju2016grad}, called Gradient-weighted Class Activation Mapping (Grad-CAM), improving upon CAM. In particular, Grad-CAM uses the class-specific gradient information flowing into the final convolutional layer of a CNN to produce a coarse localization map of the important regions in a given input image. In our research, this localization process is activated only if the class is predicted as glaucomatous. 

\section{Experiments}
\label{sec:experiment}

\textbf{Dataset: fundus images.} The total number of eye fundus images in our training dataset is 1,080, among which 220 images were used for validation purposes. Another 110 eye fundus images unseen during training and validation were used for testing purposes. These images, with the horizontal resolution varying from 1,172 to 2,500 pixels and with the vertical resolution varying from 1,500 to 3,200 pixels, have been collected at Samsung Medical Center in 2016. All images were labeled by an experienced ophthalmologist. The proportion of normal and glaucomatous images for both training and testing was 1:1 (that is, the dataset used was perfectly balanced). This study followed all guidelines for experimental investigation in human subjects, was approved by the Samsung Medical Center Institutional Review Board, and adhered to the tenets of the Declaration of Helsinki.

\textbf{Implementation.} Our approach towards glaucoma diagnosis and localization was implemented in Python 2.7, using Tensorflow 1.7 and the , pre-trained VGG16 model over ImageNet from Tensorflow Slim\footnote{https://github.com/tensorflow/models/tree/master/research/slim}. Furthermore, our approach was executed using two Intel(R) Xeon(R) E5-2620 2.4GHz CPUs and an NVIDIA GeForce GTX TITAN X GPU. Before feeding the input images to the network, they were center-cropped to remove excessive information not related to the classification task at hand (e.g., removal of a black background) and resized to $224\times224\times3$. In addition, for image normalization purposes, we changed the RGB means for our dataset as follows: 105.51, 54.52, and 16.19 (in RGB order). After changing the last layer of the pre-trained model into fully convolutional layer, the weights of all pre-trained layers were fine-tuned over our dataset. During fine-tuning, we used a batch size of 32, a learning rate of 0.0001, and data augmentation. We made use of 5-fold cross-validation for our experiments. The metrics used for evaluating the classification task are (1) accuracy, (2) recall, (3) precision, (4) F1 score, and (5) Receiver Operating Characteristic - Area Under the Curve (ROC-AUC).

\begin{figure}[htbp]
    \centering
    \subfloat[][Model error]{\includegraphics[width=0.31\textwidth]{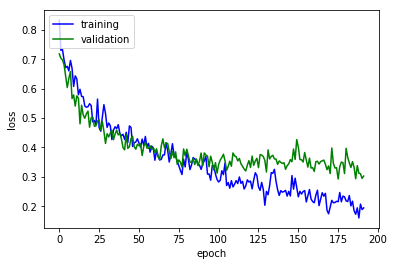}}
    \label{fig:error}
	\quad
	\subfloat[][Model accuracy]{\includegraphics[width=0.31\textwidth]{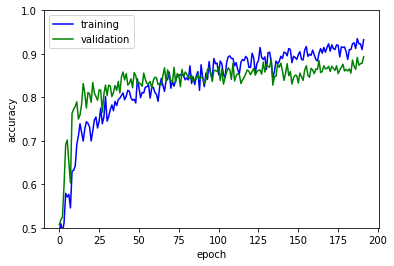}}
	\quad
	\subfloat[][ROC-AUC curve]{\includegraphics[width=0.31\textwidth]{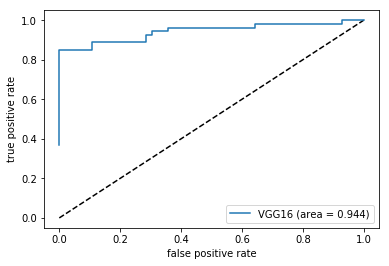}}
	\caption{Model evaluation.}
	\label{fig:result}
\end{figure}

\textbf{Experimental results.} 
Considering its effectiveness, early stopping is one of the most commonly used regularization methods in practice ~\cite{Goodfellow-et-al-2016}. Thus, as shown in Figure~\ref{fig:result}, we early stopped the training at around epoch 80, coming with a validation loss that is the lowest and that is about to rise again, even though the training loss is still decreasing over time. Apart from an accuracy of 0.91 and an ROC-AUC score of 0.94, our model achieved a recall of 0.91, a precision of 0.92, and an F1 score of 0.91.
As mentioned in Section~\ref{sec:intro}, no method is currently available that simultaneously tackles classification and localization. Moreover, a direct comparison with already existing approaches is difficult, given a lack of publicly available datasets. In Table~\ref{tab:comparison}, some information can be found about the difference in effectiveness of our approach, as obtained for the dataset made available by Samsung Medical Center, and the effectiveness of the deep learning-based approach presented in~\cite{al2017automated}, as obtained for the RIM-ONE dataset~\cite{fumero2011rim}, for the classification task only.

\begin{table*}[htbp]
\caption{Comparison with another deep learning model, using different datasets.}
\begin{center}
\begin{tabular}{|c|c|c|c|c|c|c|}
\hline
Model & Dataset & \# positives & \# negatives & Input size & Accuracy & AUC \\
\hline
\cite{al2017automated} & RIM-ONE~\cite{fumero2011rim} & 200 & 255 & $227\times227\times3$ & 0.88 & - \\
\hline
Proposed & SMC & 540 & 540 & $224\times224\times3$ & 0.91 & 0.94 \\
\hline
\end{tabular}
\label{tab:comparison}
\end{center}
\end{table*}

Apart from the use of two different datasets, thus making a direct comparison impossible, we hypothesize that some of the differences in effectiveness can be attributed to the CNN architecture used and the way data augmentation has been used. In particular,~\cite{al2017automated}~employs AlexNet, which was reported to be less effective for the ILSVRC classification task than VGG16~\cite{simonyan2014very}. Furthermore, the approach of~\cite{al2017automated} did not apply any data augmentation technique. As a result, given the limited size of the dataset used, we believe chances are high that the model of~\cite{al2017automated} did not sufficiently generalize.

\begin{figure}[htbp]
\centering
\subfloat[][Correctly localized suspicious areas.]{\includegraphics[width=0.45\linewidth]{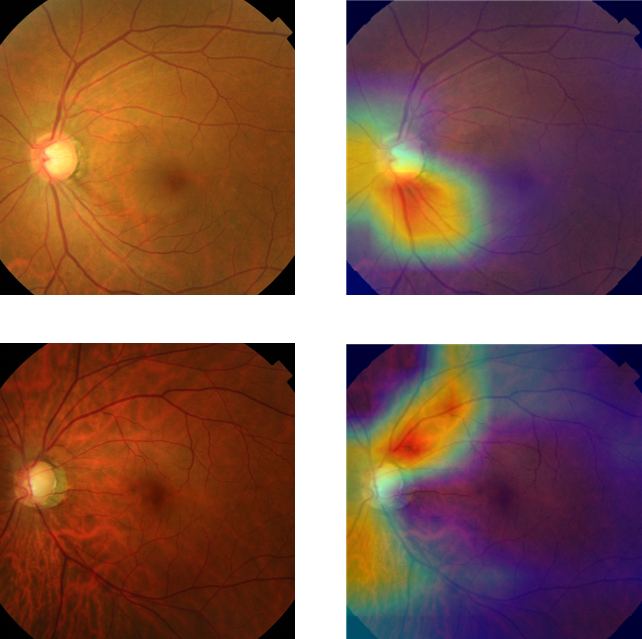}}
\qquad
\subfloat[][Wrongly localized suspicious areas.]{\includegraphics[width=0.45\linewidth]{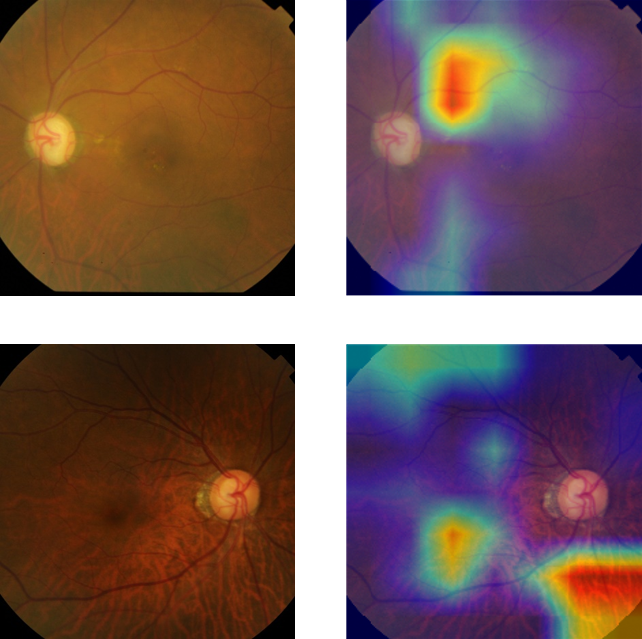}}
\caption{Localization results.}
\label{fig:localization}
\end{figure}

\begin{figure}[htbp]
\centering
\includegraphics[width=0.95\linewidth]{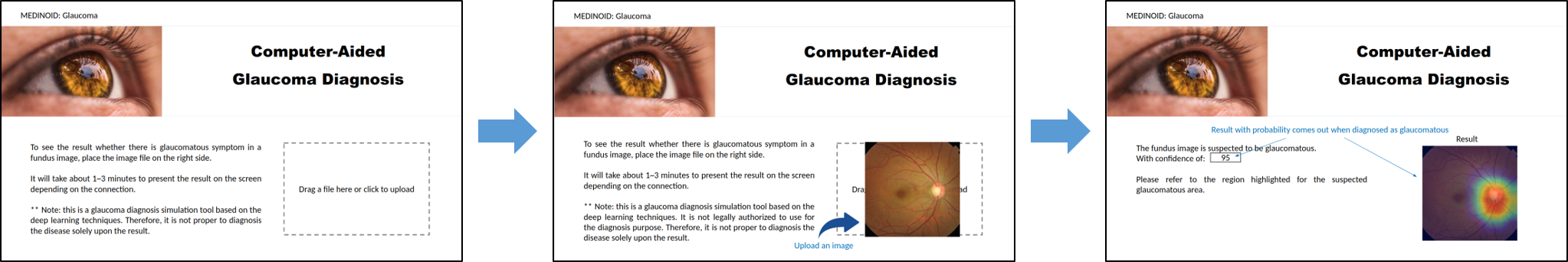}
\caption{Prototype web application for glaucoma diagnosis. Users can provide this application directly with an input image using a drag and drop approach, as shown by the figure in the middle. Furthermore, the prototype web application shows the probability of having glaucoma, highlighting the suspicious region found.}
\label{fig:web_app}
\end{figure}

Regarding localization, since this paper presents the first attempt to localize glaucoma without a segmentation ground truth, there are no proper baselines to compare against. Therefore, we asked for the opinion of a clinician at Samsung Medical Center. That way, we found that most of the glaucomatous images were correctly localized, highlighting the corresponding suspicious area, as shown in Figure~\ref{fig:localization}. However, we found that localization was also wrong for a few test examples (e.g., due to blur), thus pointing to the need for a more rigorous computational and clinical analysis. Lastly, as illustrated by Figure~\ref{fig:web_app}, we built a publicly available prototype web application that integrates our predictive model, with the goal of making glaucoma diagnosis available to a wide audience.

\section{Conclusions and Future Work}
\label{sec:conclusion}
In this paper, we introduced a predictive model for computer-aided diagnosis of glaucoma, leveraging CNNs and Grad-CAM for diagnosis and localization, respectively. Our model has been developed by making use of small-sized dataset of fundus eye images, only coming with annotations for the diagnosis task, not having annotations for the localization task. Furthermore, we integrated our model into a publicly available prototype web application, with the ultimate goal of making effective glaucoma diagnosis available to a wide audience in a user-friendly way.

\subsubsection*{Acknowledgments}

The research and development efforts described in this paper were funded by Ghent University, Ghent University Global Campus, Flanders Innovation \& Entrepreneurship (VLAIO), the Fund for Scientific Research-Flanders (FWO-Flanders), and the European Union. These efforts were also supported by the Basic Science Research Program of the National Research Foundation of Korea (NRF), funded by the Ministry of Education (grant: 2017034834).

\medskip

\bibliographystyle{ieeetr}
\bibliography{nips_2018}

\end{document}